\documentclass{article} 
\usepackage{iclr2026_conference,times}

\usepackage{amsmath,amsfonts,bm}

\def\eqref#1{equation~\ref{#1}}

\def\1{\bm{1}}

\DeclareMathAlphabet{\mathsfit}{\encodingdefault}{\sfdefault}{m}{sl}
\SetMathAlphabet{\mathsfit}{bold}{\encodingdefault}{\sfdefault}{bx}{n}

\usepackage{algorithm}
\usepackage[noEnd,commentColor=black]{algpseudocodex}
\usepackage[most]{tcolorbox}
\usepackage{hyperref}
\usepackage{url}

\usepackage{graphicx}

\usepackage[utf8]{inputenc} 
\usepackage[T1]{fontenc}    
\usepackage{hyperref}       
\usepackage{url}            
\usepackage{booktabs}       
\usepackage{amsfonts}       
\usepackage{nicefrac}       
\usepackage{microtype}      
\usepackage{xcolor}         
\usepackage{amsmath}
\usepackage{enumitem}
\usepackage{amssymb}
\usepackage[para,online,flushleft]{threeparttable}
\usepackage[nameinlink,noabbrev]{cleveref}
\usepackage{wrapfig}
\usepackage{float}

\title{Do You Really Need to Pretrain Q-Functions for Online RL Fine-Tuning? }

\author{
Perry Dong\thanks{Equal contributions. \texttt{perryd@stanford.edu} }\:\: \quad Ron Polonsky\footnotemark[1]\:\: \quad Dorsa Sadigh \quad Chelsea Finn \\
     \normalfont \ Stanford University \\
}

\definecolor{rustorange}{RGB}{204,120,47}  
\definecolor{persimmon}{RGB}{236,88,0}    
\definecolor{tomatoorange}{RGB}{224,73,42}  

\providecommand{\ours}[1][]{{\protect\color{persimmon}{IPE\textbf{#1}}}}

\definecolor{myblue}{rgb}{1,0.5,0}
\definecolor{urlteal}{HTML}{0F766E}
\usepackage{xcolor}
\definecolor{myorange}{HTML}{E67E22}
\hypersetup{
    colorlinks = true,
    linkcolor = {myorange},
    citecolor = {myorange},
    urlcolor=myorange,
    anchorcolor = black,
}

\iclrfinalcopy 
\begin{document}

\maketitle

\begin{abstract}
Pre-training followed by fine-tuning has become the dominant recipe for learning performant policies, and in value-based reinforcement learning (RL) this raises a natural question: given a pretrained policy, should the Q-function be pretrained on offline data too? Conventional wisdom suggests it should, but recent results show that online RL with a randomly-initialized Q-function can result in highly performant and reliable policies without needing to pretrain the Q-function. In this paper, we systematically study whether pretraining the Q-function actually helps when fine-tuning on top of a pretrained base policy. We find, surprisingly, that naive Q-function pretraining often provides little benefit over random initialization. We show this stems from a fundamental mismatch: the Q-function learned during pretraining targets the pretrained policy's Q-function, not the Q-function that online fine-tuning converges to, and this gap persists even after offline value maximization. Motivated by this finding, we propose \textcolor{persimmon}{I}nitialization via \textcolor{persimmon}{P}olicy \textcolor{persimmon}{E}nsemble (\ours{}), a simple method that trains multiple diverse policies and uses their pooled rollouts to bootstrap the Q-function learning in online RL. Across a suite of challenging continuous control benchmarks, \ours{} yields an average 1.26x improvement in fine-tuning performance over naive Q-function pre-training. 
\end{abstract}

\section{Introduction}
\label{sec:intro}

\begin{figure}[h]
    \vspace{-0.3cm}
  \centering
  \includegraphics[width=\linewidth]{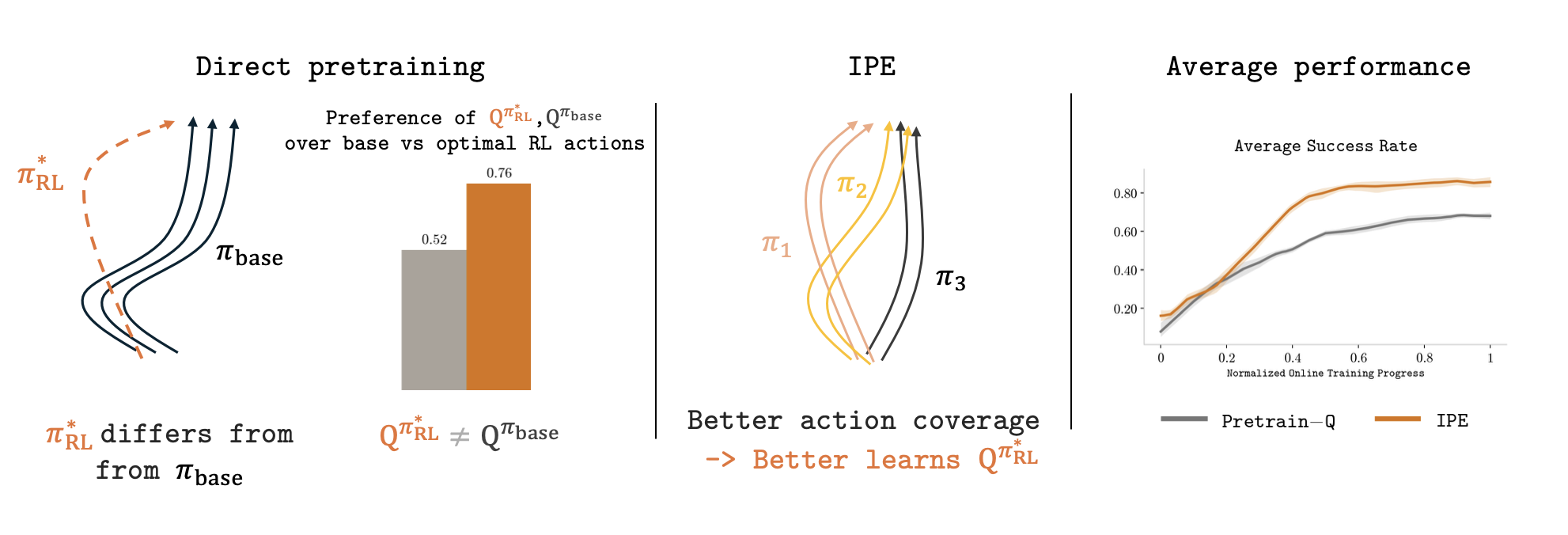}
  \caption{\footnotesize \textbf{\ours. } Left: naively pretraining does not help RL fine-tuning performance as $Q^{\pi_{\text{base}}} \neq Q^{\pi^*_{\text{RL}}}$. Mid: \ours{} enables online RL fine-tuning to better learn $Q^{\pi^*_{\text{RL}}}$ through diverse action coverage. Right: \ours{} results in an average of 26\% improvement over directly pretraining the Q-function. }
  \label{fig:pretrain_vs_no_pretrain}
\end{figure}

Pretraining followed by using reinforcement learning (RL) to finetune has become the predominant approach of learning reliable, performant policies in a range of settings from language modeling~\citep{ouyang2022traininglanguagemodelsfollow} to robotics~\citep{intelligence2025pi06vlalearnsexperience,dong2026expoftsampleefficientreinforcementlearning}. Given access to strong pretrained policies, a central question is how to best finetune them with RL. In value-based RL, in addition to the policy, the Q-function, which tells the policy which actions are good in a state, is also crucial for downstream performance. Given that the policy is pretrained on offline data, intuitively it makes sense to also pretrain the Q-function on offline data. However, recent works have shown strong results by not initializing the Q-function and only starting with a pretrained policy for online training~\citep{dong2026expoftsampleefficientreinforcementlearning}. In this work, we put this intuition to question and ask: when given a pretrained policy, should you pretrain a Q-function for online RL finetuning? 

Conventionally, offline-to-online RL~\citep{vecerik2018leveragingdemonstrations, nair2021awacacceleratingonline,nakamoto2024calqlcalibrated,ball2023efficient,dong2025expostable} works have explored jointly pretraining a Q-function and a policy, followed by online RL finetuning. However, this requires the policy to be trained jointly with the Q-function using the specific offline RL algorithm's objective, as opposed to a general pretrained policy such as with supervised learning. Adapting an existing pretrained policy to fit this joint offline RL training procedure can be challenging and costly, requiring changes to the policy's training pipeline, and its effectiveness at scale remains largely untested. Instead, we focus on the setting of only having access to a pretrained policy and data, where the task is to perform online RL finetuning and obtain the highest performance. Since the Q-function directly shapes which actions the policy is pushed toward during online RL, its initialization can substantially affect both learning speed and final performance. This suggests great care needs to be taken to examine the effects of pretraining Q-functions for downstream performance.

In this work, we answer these questions through a systematic empirical analysis to understand how Q-function pretraining affects downstream performance, and to see if there is a way to make pretraining more effective. We start by analyzing what happens when you pretrain a Q-function for online RL fine-tuning and make the surprising finding that naively pretraining the Q-function before online RL often does not result in performance improvements. We posit this is because the Q-function learned from pretraining, $Q^{\pi_{\text{base}}}$, is fundamentally different from the optimal Q-function RL fine-tuning learns, $Q^{\pi^*_{\text{RL}}}$, as evidenced by their different preferences for actions, and this results in a shift in the value landscape from pretraining to finetuning. Furthermore, we show that this difference requires online adaptation to overcome and cannot be mitigated through offline value maximization of the pretrained policy. Based on these observations, we propose a simple method to boost the performance of online RL finetuning, which we call 
\textcolor{persimmon}{I}nitialization via \textcolor{persimmon}{P}olicy \textcolor{persimmon}{E}nsemble (\ours{}). \ours{} trains multiple different policies on the same action distribution of the pretrained policy and collects rollouts with those policies to allow the Q-function to distinguish actions that the policy generates, while increasing the diversity of the data around the policy data distribution.

Our main contribution is an extensive analysis of Q-function pretraining for finetuning policies with online RL. Contrary to conventional beliefs of pretraining as much as possible, we find that directly pretraining the Q-function when starting from a pretrained policy often does not help performance of RL finetuning. On top of that, we identify the reason where Q-function pretraining is not helpful -- the offline Q-function is different than the Q-function learned by online RL. Building on these insights, we propose \ours{} for learning better Q-functions for online RL finetuning. Evaluating on complex continuous control task benchmarks, \ours{} results in an average of 26\% improvement over naively pretraining the Q-function for online RL finetuning.

\section{Related Work}

\textbf{Offline-to-Online RL.} Offline-to-online RL studies the problem of learning a policy online with interactions after training it offline. Popular approaches for offline-to-online RL involve balancing exploration~\citep{yang2023hybrid,hu2023imitation,zhang2023policyexpansion,mark2023offline}, calibrating values~\citep{nakamoto2024calqlcalibrated}, or maintaining offline data for online fine-tuning~\citep{vecerik2018leveragingdemonstrations,nair2021awacacceleratingonline,ball2023efficient,dong2025reinforcementlearningimplicitimitation,dong2025expostable}, with the goal of improving policy performance after offline RL. Offline RL methods have also been directly applied to online fine-tuning~\citep{fujimoto2019offpolicydeepreinforcementlearning,fujimoto2021minimalistapproachofflinereinforcement,hansenestruch2023idqlimplicitqlearningactorcritic,park2025flowqlearning,dong2026tqlscalingqfunctionstransformers,dong2026valueflows}. Our work differs from this line of research in that we focus on the problem of fine-tuning with online RL on top of a pretrained policy, such as a Vision-Language-Action policy trained on a diverse offline dataset using supervised learning~\citep{geminiroboticsteam2025geminirobotics15pushing,intelligence2025pi05visionlanguageactionmodelopenworld}, rather than performing the entire offline RL procedure which can be challenging. 

\textbf{RL with Pretrained Policies.} A separate line of work leverages pretrained policies as a starting point for downstream reinforcement learning, rather than relying on performing offline RL, such as jump-starting RL from a suboptimal or pretrained policy~\citep{uchendu23a}, RL without retaining offline data~\citep{zhou2025efficientonlinereinforcementlearning}, residual policy learning that trains a corrective policy on top of a fixed pretrained one~\citep{johannink2018residualreinforcementlearningrobot,silver2019residualpolicylearning}, and training BC policies for more efficient RL fine-tuning~\citep{wagenmaker2025posteriorbehavioralcloningpretraining}. Other works use on-policy methods to fine-tune pretrained policies policy~\citep{schulman2017proximalpolicyoptimizationalgorithms,schaal1996learning,rajeswaran2018learningcomplexdexterousmanipulation,ren2024diffusionpolicypolicyoptimization}. We focus on the off-policy case for increased sample-efficiency. More recently, this paradigm has been applied to fine-tuning large-scale VLA policies with online RL, where the VLA itself serves as the pretrained policy that is improved through online interaction with RL~\citep{chen2025conrftreinforcedfinetuningmethod,dong2026expoftsampleefficientreinforcementlearning}. Our work builds on this direction, and specifically investigates how pretraining Q-functions affect the performance of RL fine-tuning.

\section{Problem Formulation} 

We consider a Markov decision process (MDP) specified by the tuple $\{\mathcal{S}, \mathcal{A}, \rho, r, \gamma, T\}$, where $\mathcal{S}$ denotes the state space, $\mathcal{A}$ the action space, $\rho(s)$ the distribution over initial states, $r:\mathcal{S}\times\mathcal{A}\rightarrow\mathbb{R}$ the reward function, $\gamma\in[0,1)$ the discount factor, and $T(s'|s,a)$ the state-transition probability. The goal of RL is to find a policy $\pi$ that maximizes the expected discounted return $\mathbb{E}_{\pi}\left[\sum_{t=0}^{T}\gamma^t r(s_t, a_t)\right]$.

In this paper, we study reinforcement learning finetuning on top of a pretrained base policy $\pi_{\text{base}}$. This could, for example, be a pretrained Vision-Language-Action (VLA) model trained with supervised learning on offline data, though the experiments in this paper focus on smaller scale policies (also trained with supervised imitation) to enable extensive analysis. Concretely, we assume access to a fixed offline dataset $\mathcal{D}_{\text{offline}} = \{(s,a,r,s')\}$ of previously collected expert transitions, which is used to pretrain $\pi_\text{base}$ and, optionally, a Q-function $Q_\phi$, which is the focus of study in this work. During online finetuning, the agent collects tuples $(s_t,a_t,r_t,s_{t+1})$ through environment interaction; these are appended to a replay buffer $\mathcal{D}$ (initialized from $\mathcal{D}_\text{offline}$) and used to continue updating both the policy and the Q-function toward higher returns. Our central question is how the pretraining of $Q_\phi$ affects the sample efficiency and final performance of the resulting online RL fine-tuning.

We focus on off-policy RL, where the Q-function estimates the discounted return of the policy given a state and action, and is trained with TD learning:
\begin{equation}
    \mathcal{L}(\phi) = \mathbb{E}_{(s_t, a_{t}, s_{t+1}) \sim \mathcal{D}}\left[\left(r_t + \gamma Q_{\phi'}(s_{t+1}, \tilde{a}^*_{t+1}) - Q_\phi(s_t, a_{t})\right)^2\right],
\end{equation}
where $Q_{\phi'}$ is a target network and $\tilde{a}^*_{t+1}$ is the next action selected by the RL policy. 

\paragraph{EXPO.} We use EXPO~\citep{dong2025expostable}, a recently proposed, sample-efficient and stable RL algorithm, which has also been shown to be effective for finetuning pretrained policies such as VLA models~\citep{dong2026expoftsampleefficientreinforcementlearning}, as a general off-policy RL algorithm to analyze Q-function pretraining. EXPO maintains two parameterized policies: a base flow policy---in our case the base policy $\pi_\text{base}$, trained with a supervised loss---and a lightweight edit policy $\pi_\text{edit}$ trained to maximize the Q-function:
{\small
\begin{equation}
\begin{split}
    \mathcal{L}(\pi_\text{edit}) = -\mathbb{E}_{(s_t,a_{t})\sim\mathcal{D},\;\hat{a}_{t}\sim\pi_\text{edit}}
    \bigl[Q_\phi(s_t,\, a_{t} + \hat{a}_{t}) - \alpha \log \pi_\text{edit}(\hat{a}_{t}\mid s_t, a_{t})\bigr].
\end{split}
\end{equation}
}

The edit policy predicts an edit $\hat{a}_t$, constrained to $[-\beta, \beta]$, that is added to the base action $a_t$ to produce an edited action $\tilde{a}_t = a_t + \hat{a}_t$. This avoids backpropagating gradients through the base policy while still grounding TD updates in near-optimal actions. The final inference and TD-backup policy is an on-the-fly (OTF) policy that selects the value-maximizing candidate across $N$ sampled base and edited actions:
\begin{equation}
    \tilde{a}^* = \underset{a \;\in\; \bigcup_{i=1}^{N}\{a_i,\,\tilde{a}_i\}}{\arg\max}\; Q_\phi(s, a),
\end{equation}
with $Q_\phi$ trained via the TD objective above.

\section{Understanding the Case of Q-Function Pretraining}
\label{sec:understanding}

The predominant paradigm in machine learning pretrains a model on offline data before fine-tuning it with RL, and pretraining is typically expected to help. In online RL finetuning, this intuition would suggest that pretraining the Q-function on offline data, in addition to pretraining the policy, should give online fine-tuning a head start. In this section, we test this intuition directly. We compare online RL performance with and without Q-function pretraining, both initialized from an identical pretrained policy, and then dig into why Q-function pretraining fails to help (Section~\ref{sec:why_no_help}) even under several natural attempts to fix it.

\textbf{Setup.} We evaluate on six challenging manipulation tasks drawn from Robomimic and OGBench. The Robomimic tasks (\texttt{lift}, \texttt{can}, \texttt{square}) involve controlling a 7-DoF robot arm on tasks such as picking up a block, picking up and placing a can, and inserting a tool onto a square peg, and provide a sparse reward that only indicates task completion. The OGBench tasks provide a reward between $-n_{\text{task}}$ and $0$ depending on how many subtasks are complete: \texttt{cube} requires pick-and-place of colored blocks, \texttt{scene} requires long-horizon reasoning over interactions with multiple objects, and \texttt{puzzle} requires solving combinatorial "Lights Out" puzzles with the robot arm, testing generalization beyond the training distribution. 

Our experiments use update-to-data ratio 1 to focus on how Q-function pretraining affects downstream performance. We refer additional results on higher UTD and more algorithms to Appendix \Cref{ap:additional_experiment}. We report additional environment details in Appendix \Cref{ap:environments} and hyperparameters in Appendix \Cref{ap:hyperparameters}.

\subsection{Does Pretraining the Q-Function Help Fine-Tuning?}
\label{sec:does_it_help}

\begin{figure}[t]
  \centering
  \includegraphics[width=\linewidth]{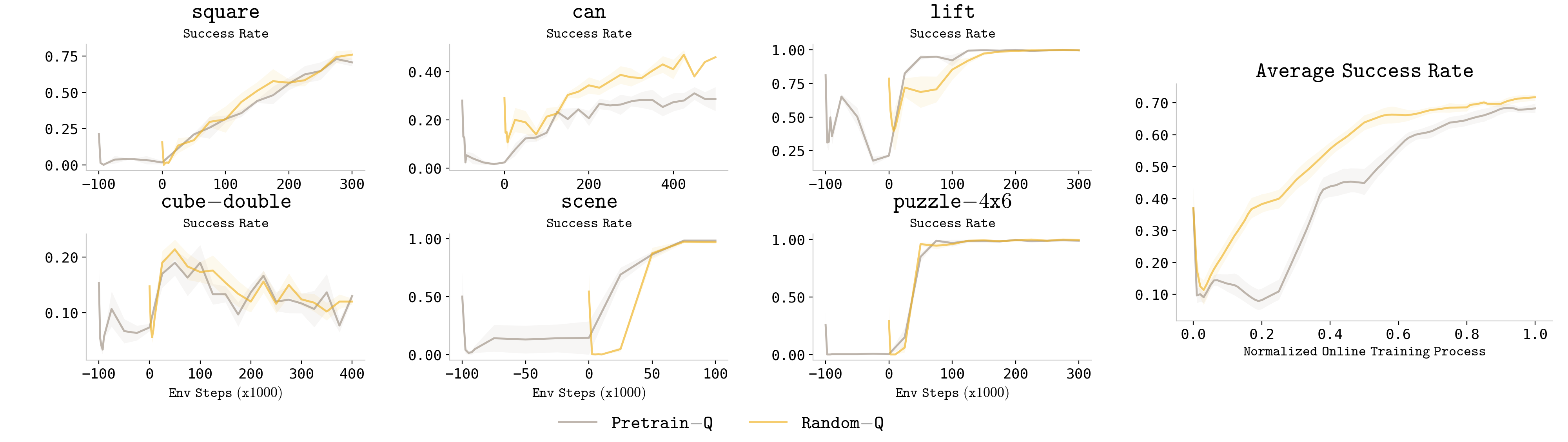}
  \caption{\footnotesize \textbf{Pretraining the Q-function does not improve performance for downstream fine-tuning.} Success rate over steps of online fine-tuning for a pretrained Q-function vs a randomly initialized Q-function, both sharing the same pretrained policy. }
  \label{fig:pretrain_vs_no_pretrain}
\end{figure}

We first ask the most direct question: starting from the same pretrained policy $\pi_{\text{base}}$, does online fine-tuning improve if the Q-function is also pretrained via offline TD learning with $\mathcal{D}_{\text{offline}}$, compared to initializing the Q-function from scratch and learning it entirely online? We pretrain the policy using supervised learning as a stable loss for pretraining. Figure~\ref{fig:pretrain_vs_no_pretrain} compares these two pipelines across all six tasks.

Contrary to the intuition that pretraining should help, we find that pretraining the Q-function provides little to no improvement in downstream fine-tuning performance, and in several tasks it actually hurts online learning relative to fine-tuning on top of a randomly-initialized Q-function. Even in the cases that show a modest acceleration early in online training, this advantage disappears as fine-tuning progresses, with both pipelines converging to similar performance. This is surprising: instead of reducing the burden of online TD learning, pretraining the Q-function did not improve performance. We investigate why this happens below.

\begin{tcolorbox}[colback=white, colframe=myorange, boxrule=0.5pt, arc=2pt, left=6pt, right=6pt, top=4pt, bottom=4pt] \textcolor{myorange}{Takeaway 1: Naive Q-function pretraining does not accelerate fine-tuning.}
Despite pretraining on offline data, the Q-function performs no better --- and sometimes worse --- than a randomly-initialized Q-function once online fine-tuning begins.
\end{tcolorbox}

\subsection{Why is Q-Function Pretraining Not Effective?}
\label{sec:why_no_help}

To understand why pretraining the Q-function fails to translate into a fine-tuning advantage, we examine how well the pretrained Q-function actually estimates the value of good actions. We introduce a \emph{preference accuracy} metric: the fraction of state-action comparisons in which the learned Q-function prefers an action from $\pi^*$ over an action from $\pi_{\text{base}}$, where $\pi^*$ is an estimate of the optimal policy (see Appendix~\ref{ap:preference} for details). A well-calibrated Q-function should generally prefer $\pi^*$'s actions.

\begin{figure}[t]
  \centering
  \includegraphics[width=\linewidth]{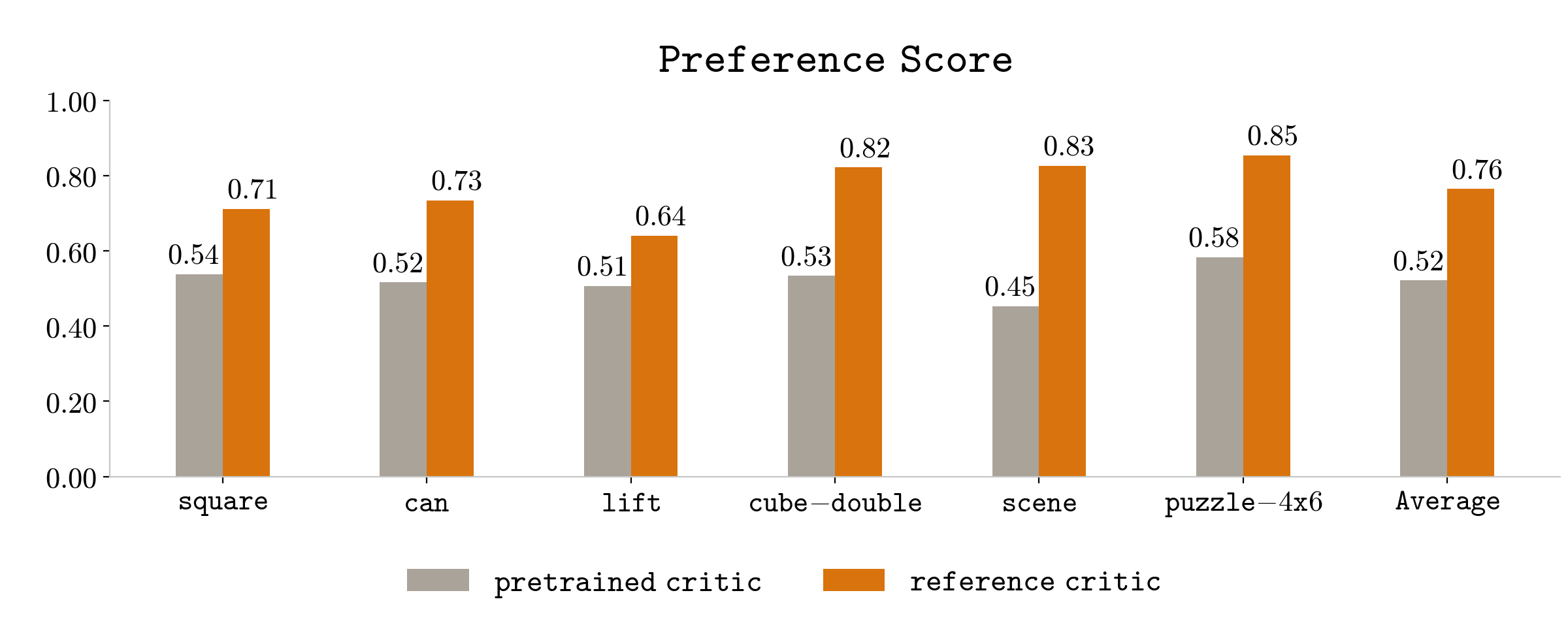}
  \caption{\footnotesize \textbf{Preference of $Q^{\pi_{\text{base}}}$ and $Q^{\pi^*_{\text{RL}}}$.} $Q^{\pi_{\text{base}}}$ is not the same as $Q^{\pi^*_{\text{RL}}}$ even for a near-optimal base policy, but the Q-functions have preferences close to each other.   }
  \label{fig:bc_vs_rl}
\end{figure}

\textbf{Why does pretraining not help?} We hypothesize that the central issue is a mismatch between $Q^{\pi_{\text{base}}}$, the value function the offline data supports, and $Q^{\pi^*_{\text{RL}}}$, the value function that fine-tuning is learning. To isolate this gap from confounds due to a weak base policy, we train a base policy to near-100\% success rate and pretrain its Q-function to convergence --- as close to $Q^{\pi_{\text{base}}}$ for an optimal distribution of the base policy as offline pretraining can get. We emphasize that, in practice, it is rare for a base data distribution to be optimal enough to support training a policy offline to 100\% performance, which is precisely why we construct this setting to characterize in the limit what the pretrained Q-function converges to. We then use the preference accuracy metric to compare how the learned $Q^{\pi_{\text{base}}}$ and $Q^{\pi^*_{\text{RL}}}$ each rank actions sampled from the base policy $\pi_{\text{base}}$ against actions sampled from $\pi^*_{\text{RL}}$, where $\pi^*_{\text{RL}}$ is a ground truth RL policy trained to 100\% performance. We present the results in~\Cref{fig:bc_vs_rl}. 

Two observations emerge. First, $Q^{\pi_{\text{base}}} \neq Q^{\pi^*_{\text{RL}}}$: even with a near-optimal base policy, the offline-pretrained critic does not coincide with the critic that online fine-tuning ultimately aims to learn, so $Q^{\pi^*_{\text{RL}}}$ must be learned through online interaction regardless of the optimality of the base distribution. Second, despite this mismatch, the two value functions are not wildly different as the preference of the two Q-functions are quite close, suggesting a nontrivial fraction of optimal actions learned by RL belongs within the distribution of $\pi_{\text{base}}$.

\begin{tcolorbox}[colback=white, colframe=myorange, boxrule=0.5pt, arc=2pt, left=6pt, right=6pt, top=4pt, bottom=4pt]
\textcolor{myorange}{Takeaway 2: The pretrained critic targets the wrong value function.}
$Q^{\pi_{\text{base}}}$ is not the same as $Q^{\pi^*_{\text{RL}}}$ --- the Q-function learned through pretraining is different from the optimal Q-function learned with online RL finetuning --- but the two are also not entirely unrelated, since a meaningful share of actions from the pretrained policy remain optimal after fine-tuning.
\end{tcolorbox}

\begin{figure}[h]
  \centering
  \includegraphics[width=\linewidth]{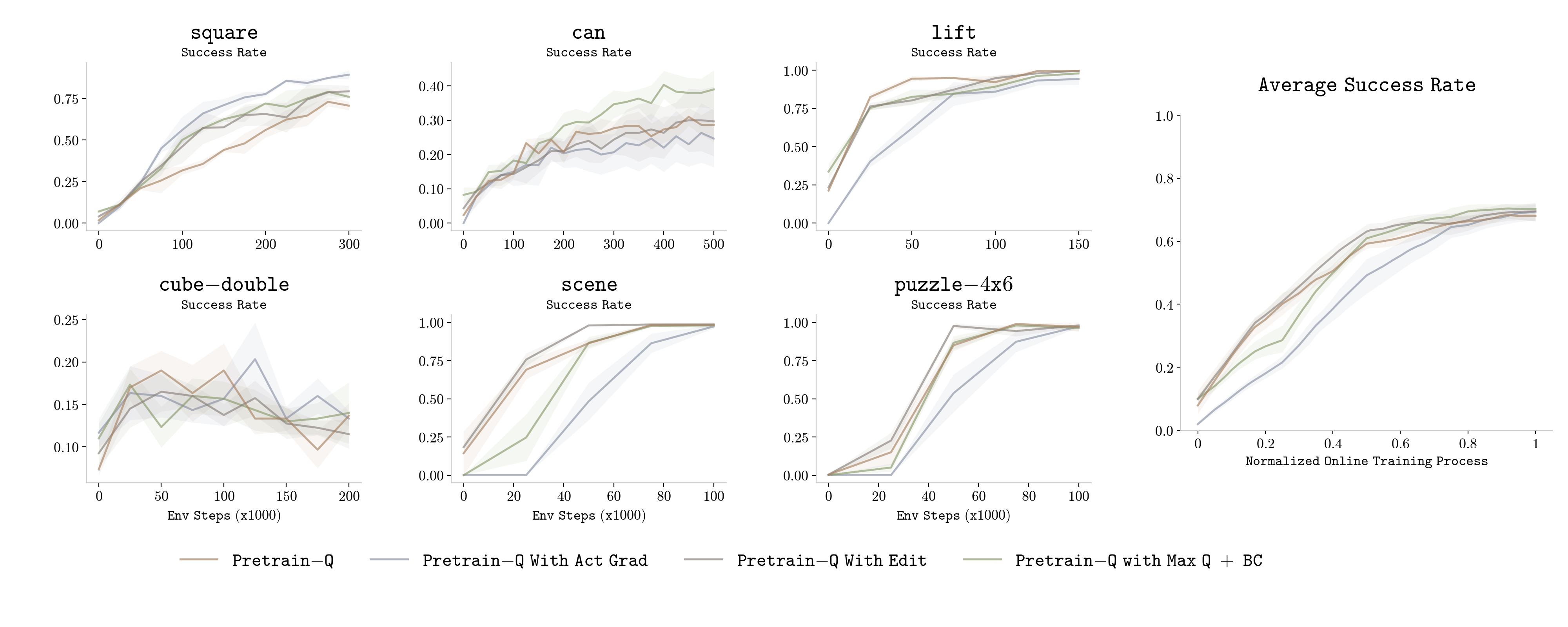}
  \caption{\footnotesize \textbf{Performance of online RL after offline value maximization.} Performing offline value maximization does not improve the performance over directly pretraining, suggesting online data and learning is necessary. }
  \label{fig:value_max}
\end{figure}

\begin{figure}[h]
  \centering
  \includegraphics[width=\linewidth]{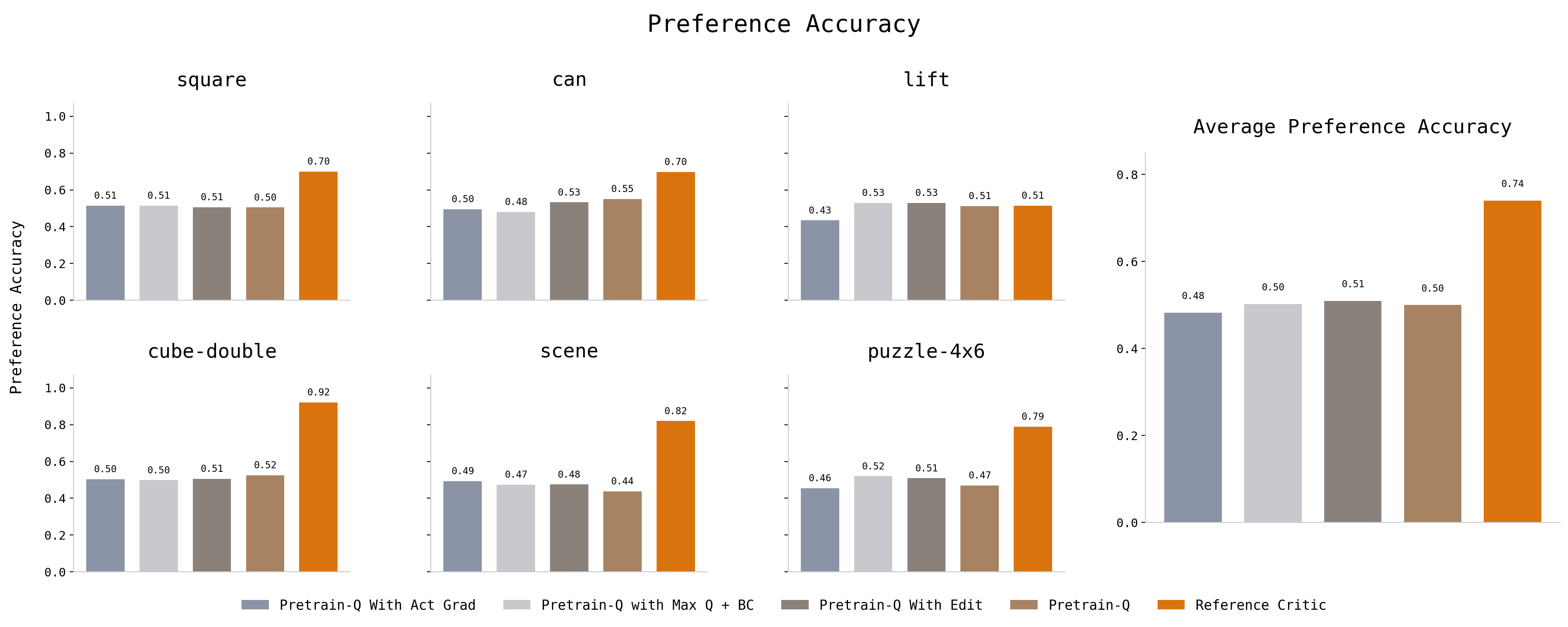}
  \caption{\footnotesize \textbf{Preference score analysis of offline value maximization.} The preference of actions after performing offline value maximization stays similar to pretraining without value maximization. }
  \label{fig:vb}
\end{figure}

\textbf{Does offline value maximization close the gap?} Since $Q^{\pi_{\text{base}}} \neq Q^{\pi^*_{\text{RL}}}$, a natural fix is to perform value maximization \emph{during} pretraining, rather than simply fitting $Q^{\pi_{\text{base}}}$, in the hope of pushing the pretrained critic closer to $Q^{\pi^*_{\text{RL}}}$ before online fine-tuning even starts. We test three such approaches: (1) learning the edit policy from EXPO, (2) distilling a separate one-step policy that maximizes Q-value on top of the pretrained policy (max Q + BC loss) similar to \citet{park2025flowqlearning}, and (3) applying action gradients directly to the pretrained policy alongside value maximization. We see in ~\Cref{fig:value_max} and ~\Cref{fig:vb} across all three, neither final performance nor preference accuracy improves over naive Q-pretraining, indicating that offline value maximization alone is not sufficient to bridge the gap identified above and online data is needed.

\begin{tcolorbox}[colback=white, colframe=myorange, boxrule=0.5pt, arc=2pt, left=6pt, right=6pt, top=4pt, bottom=4pt]
\textcolor{myorange}{Takeaway 3: Fixing pretraining post hoc through offline value maximization is insufficient.}
Offline value-maximization techniques do not close the gap between $Q^{\pi_{\text{base}}}$ and $Q^{\pi^*_{\text{RL}}}$; more data helps incrementally but does not resolve the underlying mismatch.
\end{tcolorbox}

Taken together, these results indicate that naively pretraining the Q-function on offline data is not an effective way to obtain better performance for fine-tuning: pretraining converges to the wrong target and offline value maximization during pretraining is not enough to fix this. Instead, the Q-function online fine-tuning learns must be acquired through online interaction. This motivates a different way of using the offline data to improve performance for online RL fine-tuning, which we introduce next.

\section{\ours{}: Initialization via Policy Ensemble}
\label{sec:method}

\begin{figure}[t]
  \centering
  \includegraphics[width=\linewidth]{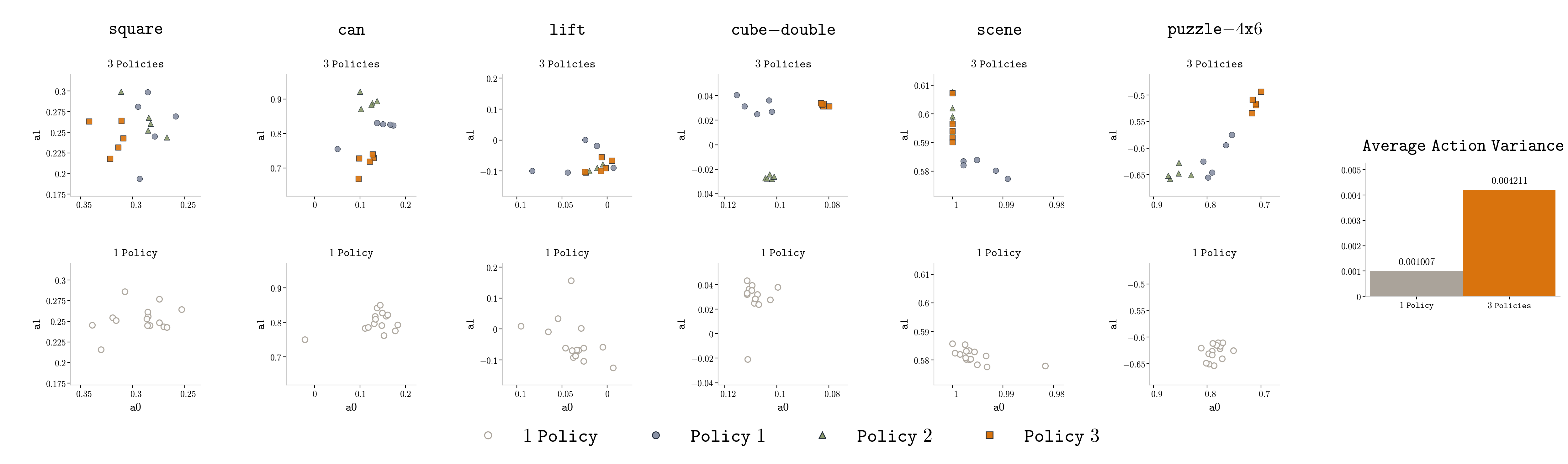}
  \caption{\footnotesize \textbf{Actions from a single policy vs multiple policies.} Actions sampled from one pretrained policy are significantly less diverse than actions pooled across three different policies trained on the same action distribution, the latter giving the Q-function meaningful coverage to learn from.}
  \label{fig:diverse}
\end{figure}

\begin{figure}[h]
  \centering
  \includegraphics[width=\linewidth]{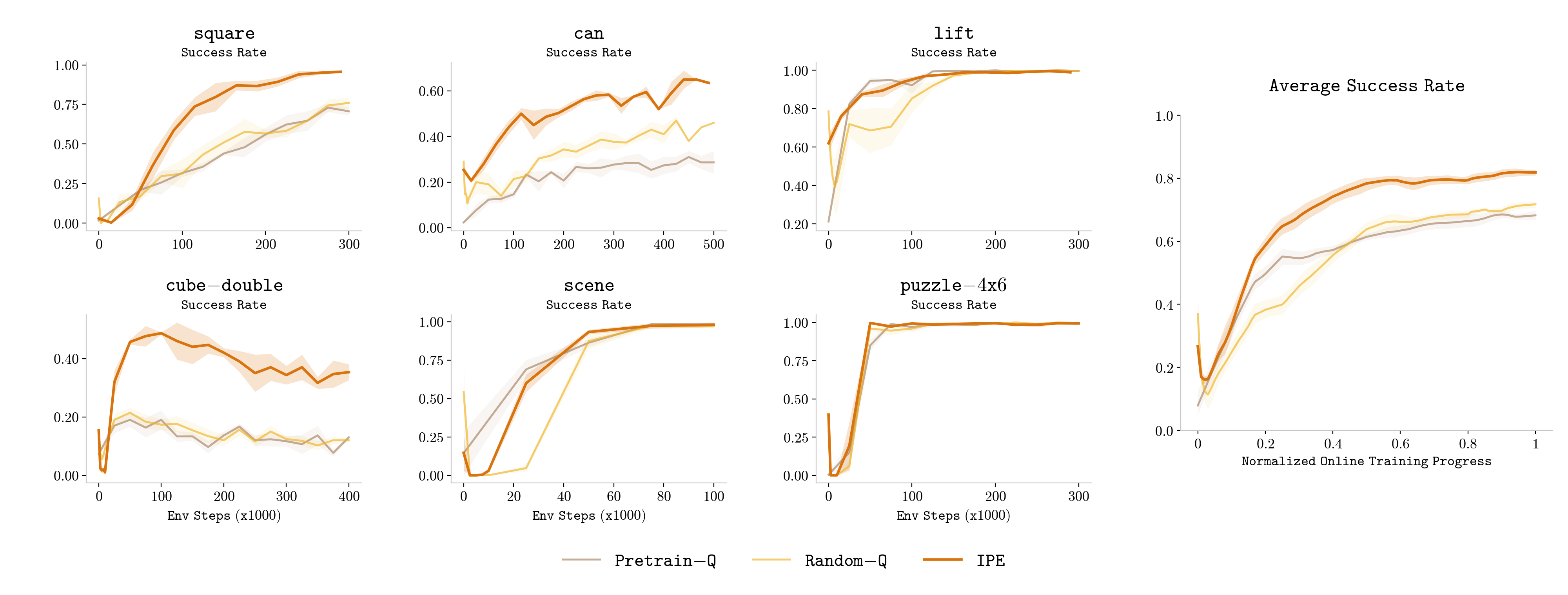}
  \caption{\footnotesize \textbf{\ours{}.} Through using multiple policies to collect diverse rollouts, \ours{} is able to better learn $Q^{\pi^*_{\text{RL}}}$ from RL finetuning.}
  \label{fig:against_pretraining}
\end{figure}

So far, we have seen that naively pretraining the Q-function on offline data does not meaningfully improve fine-tuning performance, because pretraining converges to $Q^{\pi_{\text{base}}}$ which is different to the $Q^{\pi^*_{\text{RL}}}$ that online fine-tuning actually needs (Section~\ref{sec:why_no_help}). This raises the question: is there a way to make better use of the offline data during pretraining, one that enables online RL finetuning to better learn  $Q^{\pi^*_{\text{RL}}}$?

\textbf{Key idea.} The offline dataset encodes a prior over what a near-optimal policy would predict, and Section~\ref{sec:why_no_help} showed that this prior is not far from $Q^{\pi^*_{\text{RL}}}$. The failure of direct pretraining is not that the data is uninformative, but that pretraining directly on $\pi_{\text{base}}$ overly constrains the Q-function to $Q^{\pi_{\text{base}}}$. This problem is compounded by a second issue: a single trained policy, even an expressive one such as a diffusion or flow-matching policy, tends to produce a narrow action distribution. When the Q-function is pretrained against actions from only one such policy, it receives almost no coverage over alternative actions at a given state, and the resulting $Q^{\pi_{\text{base}}}$ collapses toward $V^{\pi_{\text{base}}}$ rather than capturing meaningful action-value structure (\Cref{sec:diversity}).
\begin{wrapfigure}{r}{0.485\textwidth}
    \begin{minipage}[t]{0.485\textwidth}
    \begin{algorithm}[H]
      \caption{\ours{}}\label{alg:ours_pretrain}
      \begin{algorithmic}[1]
        \Require Offline data $\mathcal{D}_{\text{offline}}$, base policy $\pi_{\text{base}}$
        \State Train $N$ policies $\{\pi_i\}_{i=1}^N$ via supervised learning on the $\pi_{\text{base}}$ action distribution 
        \State Initialize replay buffer $\mathcal{D} \leftarrow \emptyset$
        \For{each policy $\pi \in \{\pi_{\text{base}}, \pi_1, \dots, \pi_N\}$}
          \State Roll out $\pi$ in the environment to collect transitions $\{(s, a, r, s')\}$
          \State $\mathcal{D} \leftarrow \mathcal{D} \cup \{(s, a, r, s')\}$
        \EndFor
        \State \Return replay buffer $\mathcal{D}$ for online RL
      \end{algorithmic}
      \label{alg:ours}
    \end{algorithm}
  \end{minipage}
\end{wrapfigure}

Our key idea is to break this narrowness by initializing the Q-function on rollouts from a \emph{set of different policies}, all trained on the same underlying data distribution as $\pi_{\text{base}}$. Because each policy is a different function approximator (or trained with different seeds/subsets of data), their rollouts are different from one another even though they target the same behavior distribution, giving the Q-function meaningful coverage over alternative actions at each state. This diversified coverage can allow the Q-function to not be constrained to $Q^{\pi_{\text{base}}}$ and get closer to $Q^{\pi^*_{\text{RL}}}$ during RL fine-tuning. 

\textbf{\ours{} algorithm.} \ours{} instantiates this idea in three steps: (1) obtain/pretrain a base policy $\pi_{\text{base}}$ on the offline dataset; (2) train a set of $N$ additional policies on the same action distribution that $\pi_{\text{base}}$ was trained on --- this can use the identical dataset or different subsets thereof; (3) roll out each of the $N{+}1$ policies in the environment and use the resulting transitions to bootstrap the replay buffer. At $N=0$, this recovers the warm up phase for online RL. As we show in ~\Cref{sec:scaling}, larger $N$ typically yields higher performance. We summarize \ours{} in Algorithm~\ref{alg:ours}.

\section{Experiments}
\label{sec:experiments}

Our experiments study how well \ours{} enables effective fine-tuning on top of a pretrained base policy. Concretely, we investigate the following questions: (1) Does a single pretrained policy in fact produce narrow, low-diversity action rollouts? (2) Does \ours{} improve over both naive Q-pretraining and no pretraining at all? and (3) How does performance scale with the number of policies $N$ used for \ours{}?

\subsection{Does a Single Policy Produce Narrow Action Coverage?}
\label{sec:diversity}

\begin{wrapfigure}[16]{R}{0.38\textwidth}
    \vspace{-0.58cm}
    \centering
    \includegraphics[width=0.4\columnwidth]{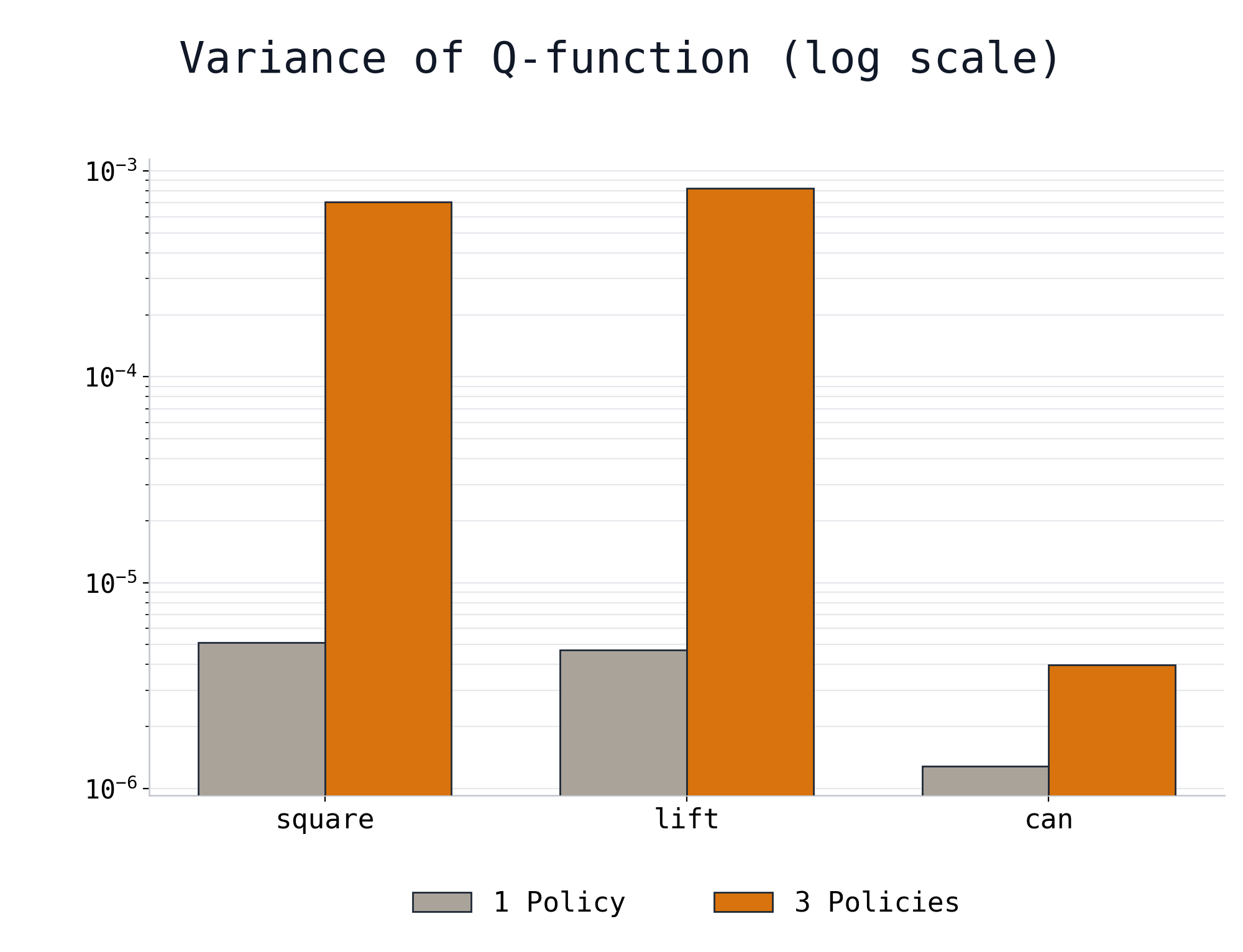}
    \caption{
        \footnotesize 
        \textbf{Variance of Q-functions. } The Q-function trained on data from 1 policy exhibits very low variance, where the Q-function is essentially deterministic.}
    \vspace{-0.9cm}
    \label{fig:log_scale}
\end{wrapfigure}

We first test the premise behind \ours{}: that a single pretrained policy yields insufficiently diverse actions for the Q-function to learn beyond $V^{\pi_{\text{base}}}$. Figure~\ref{fig:diverse} visualizes the actions sampled from a single pretrained policy alongside actions sampled from three independently-trained policies on the same data. We find that actions from the single policy cluster tightly, exhibiting substantially less diversity than the pooled actions from three separate policies, which spread more broadly over the action space despite all three being trained on identical data. This confirms that a single policy's rollouts provide limited action coverage for Q-function pretraining. Further, we show the variance across actions of Q-functions pretrained on data from 1 policy vs 3 policies in \Cref{fig:log_scale}. The Q-function trained on rollouts from 1 policy exhibits very low variance indicating that it is deterministic. Pretraining the Q-function against rollouts from multiple policies, rather than a single policy, gives the critic coverage over alternative actions and helps it escape the $Q^{\pi_{\text{base}}} \approx V^{\pi_{\text{base}}}$ collapse.

\subsection{Does \ours{} Improve Over Pretraining and No Pretraining?}
\label{sec:against_pretraining}

Next, we compare \ours{} against the two approaches from Section~\ref{sec:does_it_help}: fine-tuning with a naively-pretrained Q-function, and fine-tuning with a randomly-initialized Q-function, all of which of trained using the same finetuning procedure on the same pretrained policy, to evaluate the effectiveness of \ours{} in improving online RL fine-tuning performance.  Figure~\ref{fig:against_pretraining} reports success rate over the course of fine-tuning across our six tasks, along with the preference accuracy metric from Section~\ref{sec:why_no_help}. We find that \ours{} outperforms both baselines in fine-tuning to higher success rates, indicating its effectiveness to enable RL fine-tuning to better learn $Q^{\pi^*_{\text{RL}}}$ and obtain higher performance.

\subsection{Scaling with the Number of Policies}
\label{sec:scaling}

\begin{figure}[t]
  \centering
  \includegraphics[width=\linewidth]{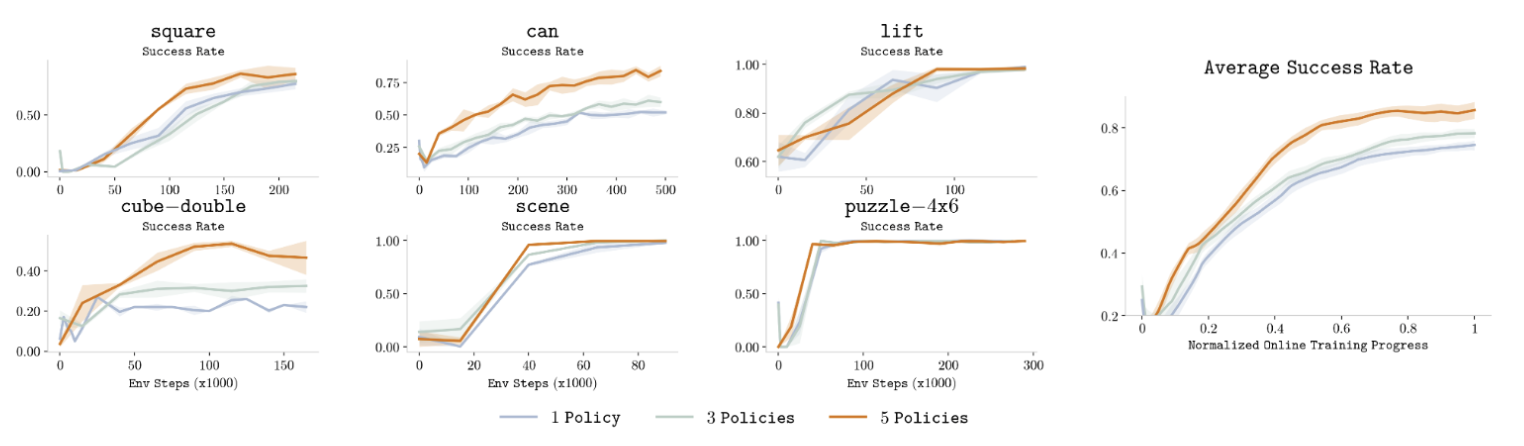}
  \caption{\footnotesize \textbf{Scaling the number of policies for \ours{}.} Increasing the number of policies used to collect rollouts results in better performance.}
  \label{fig:scaling}
\end{figure}

Finally, we study how \ours{}'s performance scales with the number of policies $N$ used to collect pretraining data. To do so, we vary $N$ across a range of values while holding the total number of pretraining trajectories fixed, so that increasing $N$ corresponds to collecting data from a more diverse set of behavior policies rather than simply having more data. The fine-tuning performance is shown in \Cref{fig:scaling}. We observe a general trend of improving fine-tuning performance as $N$ increases, with $N=5$ having the best performance in our experiments. This suggest the broader coverage provided by \ours{} that enables the policy prior to have a wider range of behaviors and state-action combinations than any single policy would provide on its own is a simple and effective method for improving downstream fine-tuning performance.

\section{Discussion} \label{sec:discussion}

In this work, we present a systematic study of Q-function pretraining for reinforcement learning finetuning on top of pretrained policies, and \ours{}, a simple method for learning more effective Q-functions for online RL finetuning. Contrary to the conventional intuition that more pretraining is always beneficial, we find that naively pretraining the Q-function on offline data \emph{does not} translate into improved downstream finetuning performance. We trace this behavior to a fundamental mismatch between the Q-function learned during offline pretraining and the one required for online RL. We show that \ours{} improves significantly over naive Q-function pretraining. Despite these results, \ours{} has limitations. Our analysis focuses specifically on finetuning with off-policy RL. While we expect the underlying mismatch we identify to generalize, examining similar trends for on-policy methods remains an important direction for future work. Additionally, \ours{}'s benefit comes from collecting rollouts across multiple policies, which introduces additional data-collection overhead relative to pretraining a Q-function on a single fixed policy's data. We leave the problem of reducing this overhead as future work.

\bibliography{iclr2026_conference}
\bibliographystyle{iclr2026_conference}

\appendix

\clearpage

\section{Additional Experiments}
\label{ap:additional_experiment}

To further examine the effect of pretraining the Q-function on downstream performance, we report results for EXPO and RLPD with a UTD ratio of 20. The setup mirrors the experiments in \Cref{fig:pretrain_vs_no_pretrain}: we are given a pretrained policy, and the Q-function is pretrained on top of it. In the non-pretrained variant, the Q-function is instead initialized randomly and trained solely online. Across all settings, we find that pretraining the Q-function does not translate into a meaningful improvement in downstream performance for either method, and in some cases yields slightly lower performance. As discussed in \Cref{sec:why_no_help}, the Q-function learned during pretraining differs fundamentally from the one learned by online RL; as a result, pretraining on offline data fails to help compared to a randomly initialized Q-function.

\begin{figure}[h]
  \centering
  \includegraphics[width=\linewidth]{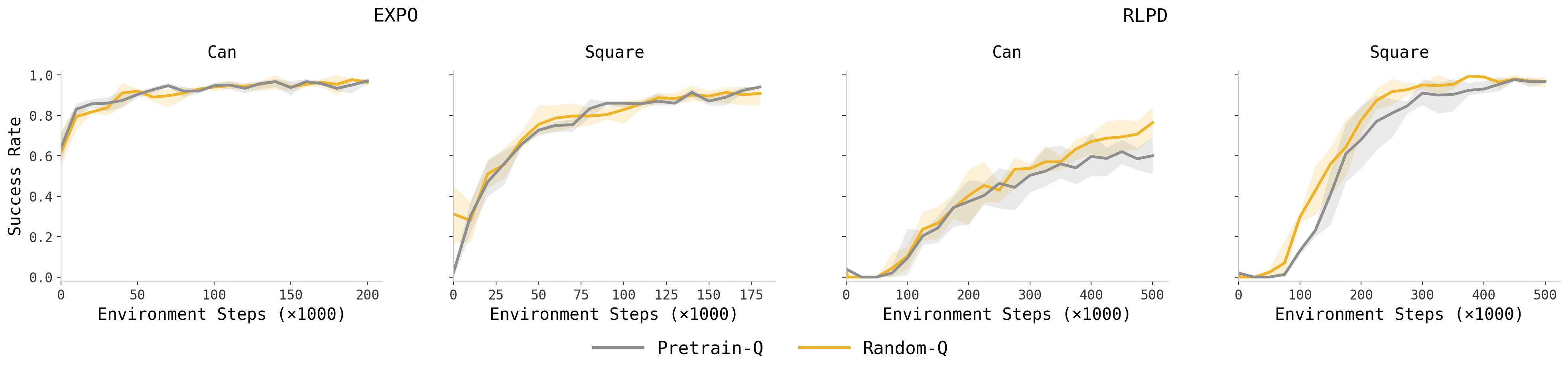}
  \caption{\footnotesize \textbf{Evaluation success rates of pretraining versus no pretraining of Q-functions.} Pretraining Q-functions for both EXPO with UTD 20 and RLPD with UTD 20 do not result in improvement in performance compared to a randomly initialized Q-function.  }
  \label{fig:additional}
\end{figure}

\section{Experiment Details}

\subsection{Preference Accuracy} \label{ap:preference}

We use a \emph{preference accuracy} metric to evaluate whether a learned critic recovers action preferences identified by a strong reference critic. The reference critics are trained RL agents that successfully solve the task to 100\% or close to 100\% performance, which we can view as $Q^{\pi^*_{\text{RL}}}$. To reduce variance from any single trained critic and recover the distribution of $Q^{\pi^*_{\text{RL}}}$, we train $10$ reference critics and average over them. For each state $s_i$, we compare two actions: an action $a_i^{\mathrm{base}}$ from the base policy and a reference action $a_i^{\mathrm{ref}}$. The preference accuracy evaluates how often a critic prefers a reference action $a_i^{\mathrm{ref}}$ over a base action $a_i^{\mathrm{base}}$. A Q-function that is calibrated should generally prefer $a_i^{\mathrm{ref}}$ over $a_i^{\mathrm{base}}$.

\subsection{Hyperparameters }
\label{ap:hyperparameters}
This section provides additional details on our training procedure, pretraining datasets, and actor-generated rollout configurations across all six environments. Each row reports ($N_{\mathrm{add}}$), the number of additional generated rollouts, and ($p_f$), the fraction of failed trajectories among those rollouts. Within each environment, the dataset size and ($N_{\mathrm{add}}$) are held fixed across settings.

\begin{table*}[h]
\centering
\caption{Policy generated rollouts configurations.}
\label{tab:all_environment_actor_data}
\vspace{0.05cm} 
\renewcommand{\arraystretch}{1.2}
\scriptsize
\begin{tabular}{lcccccc}
\hline
\textbf{Setting} & \textbf{square (ph)} & \textbf{can (mh)} & \textbf{lift (ph)} & \textbf{cube-double } & \textbf{scene} & \textbf{puzzle-4x6} \\
\hline\\[-2.5ex]
Pretraining demonstrations & $100$ & $10$ & $10$ & $40$ & $50$ & $10$ \\
$N_{\mathrm{add}},\quad p_f$ & $200,0.3$ & $20,0.1$ & $50,0.1$ & $80,0.1$ & $100,0.1$ & $30,0.1$ \\[0.5ex]
\hline
\end{tabular}
\end{table*}
\clearpage

\begin{table*}[!t]
\centering
\caption{Shared EXPO model and optimization hyperparameters.}
\label{tab:shared_hyperparameters}
\vspace{0.05cm} 
\renewcommand{\arraystretch}{1.2}
\scriptsize
\begin{tabular}{p{0.58\textwidth} p{0.32\textwidth}}
\hline
\textbf{Hyperparameter} & \textbf{Value} \\
\hline\\[-2.5ex]
Base actor optimizer & AdamW \\
Edit actor optimizer & Adam \\
Critic optimizer & Adam \\
Temperature optimizer & Adam \\
Actor learning rate & $3 \times 10^{-4}$ \\
Critic learning rate & $3 \times 10^{-4}$ \\
Adam coefficients $(\beta_1,\beta_2,\epsilon)$ & $(0.9,0.999,10^{-8})$ \\
Discount factor $\gamma$ & $0.99$ \\
Critic target update rate & $0.005$ \\
Actor target update rate & $0.001$ \\
Update-to-data ratio & $1$ \\
Batch size & $256$ \\
Offline-to-online batch ratio & $0.5$ \\
Critic hidden dimensions & $(256,256,256)$ \\
Number of Q-functions & $10$ \\
Number of min Qs & $2$ \\
Critic layer normalization & True \\
Diffusion denoising steps & $10$ \\
Diffusion noise schedule & Variance preserving \\
Diffusion time-embedding dimension & $128$ \\
Diffusion residual blocks & $3$ \\
Diffusion actor layer normalization & True \\
Number of action samples & $8$ \\
Edit action scale & $0.1$ \\[0.5ex]
\hline
\end{tabular}
\end{table*}

\begin{table*}[!t]
\centering
\caption{Additional hyperparameters used for the Max Q + BC and Action-gradient Q-pretraining variants.}
\label{tab:additional_hyperparameters}
\vspace{0.05cm} 
\renewcommand{\arraystretch}{1.2}
\scriptsize
\begin{tabular}{p{0.25\columnwidth} p{0.42\columnwidth} p{0.23\columnwidth}}
\hline
\textbf{Variant} & \textbf{Hyperparameter} & \textbf{Value} \\
\hline\\[-2.5ex]
Max Q + BC & Offline pretraining updates & $100{,}000$ \\
Max Q + BC & Distillation coefficient $\alpha_{\mathrm{FQL}}$ & $10.0$ \\
Q Action gradient & Offline pretraining updates & $100{,}000$ \\
Q Action gradient & Q-gradient coefficient & $0.05$ \\
Q Action gradient & Gradient clipping threshold & $1.0$ \\[0.5ex]
\hline
\end{tabular}
\end{table*}

\subsection{Environments}
\label{ap:environments} 
Our evaluations use a set of tasks from  OGBench~\citep{park2025ogbenchbenchmarkingofflinegoalconditioned} and Robomimic~\citep{mandlekar2021matterslearningofflinehuman}. OGBench is a large-scale benchmark designed for goal-conditioned reinforcement learning (RL), where the agent must learn to reach any state from any other state in the dataset using as few steps as possible. In this work, we focus on a diverse set of manipulation tasks that evaluate the agent's object manipulation, sequential reasoning, and combinatorial generalization capabilities. Each task is compatible with  the standard reward-maximizing RL algorithms and its reward ranges from $-n_{\text{tasks}}$ to $0$, depending on the number of completed subtasks.
We use the following OGBench datasets for each domain:
\begin{itemize}
    \item \texttt{cube-double-singletask-task2-v0}
    \item \texttt{scene-singletask-task2-v0}
    \item \texttt{puzzle-4x6-singletask-task2-v0}

\end{itemize}

The \texttt{cube-double} family of tasks involves complex pick-and-place manipulation of multiple colored cube blocks, where the robot arm must arrange cubes into designated target configurations. The \texttt{scene} tasks require long-horizon reasoning and interaction with diverse objects in unstructured environments, where the agent must sequentially combine learned manipulation primitives, such as opening and closing drawers or windows. The \texttt{puzzle} task tests the generalization capabilities of the agent, requiring it to achieve a desired color configuration of a puzzle. All experiments follow the official evaluation protocols and metrics defined by OGBench~\citep{park2025ogbenchbenchmarkingofflinegoalconditioned}. For each setting, we report the mean and standard deviation of the success rate over three random seeds.

For Robomimic, we evaluate on the \texttt{lift-ph}, \texttt{square-ph}, and  \texttt{can-mh} tasks, which require lifting a block, transferring a can into a target bin, and inserting a nut onto a square peg, respectively. The offline datasets are collected through RoboTurk~\citep{mandlekar2018roboturkcrowdsourcingplatformrobotic}, a remote teleoperation platform, and consist entirely of successful demonstrations. The PH datasets contain demonstrations collected by a single experienced teleoperator, whereas the MH datasets consist of demonstrations collected by six teleoperators with varying levels of expertise (two experienced, two intermediate, and two novice operators).

\end{document}